\documentclass[10pt,letterpaper]{article}
\usepackage{arxiv}
\usepackage[T1]{fontenc}
\usepackage[utf8]{inputenc}
\usepackage{microtype}
\usepackage{amsmath,amssymb,amsthm}
\usepackage{graphicx}
\usepackage{float}
\usepackage{tikz}
\usetikzlibrary{arrows.meta}
\usepackage{booktabs}
\usepackage{array}
\usepackage{xcolor}
\usepackage{listings}
\usepackage[numbers,sort&compress]{natbib}
\usepackage{hyperref}
\hypersetup{
  colorlinks=true,
  linkcolor=blue!45!black,
  citecolor=green!35!black,
  urlcolor=blue!50!black,
  pdfauthor={Shenghao Ding},
  pdftitle={JET: Justification Evaluation in Transformer}
}
\renewcommand{\undertitle}{}
\renewcommand{\headeright}{}
\title{JET: Justification Evaluation in Transformer}
\author{Shenghao Ding\\Yet Another AI\\
\texttt{shenghao.ding@yetanother.ai}}
\date{}

\begin{document}
\maketitle
\begin{abstract}
JET uses pretrained language and vision-language models to select among a finite set of answers
without requiring additional training. It evaluates candidate likelihoods directly and
shares computation across candidates. Experiments on consumer
GPUs assess decision accuracy and execution cost. Full-set measurements
on an Arc A770 with Vulkan and an RTX 4090 with Vulkan and CUDA cover all
14,042 MMLU test questions. In the RTX 4090 Vulkan series, Qwen3.5-27B reaches
85.09\% at 2.21 requests/s, while Qwen3.6-35B-A3B reaches 82.62\% at
4.17 requests/s.
The accuracy--throughput comparison covers model and hardware choices, with
Jev as an external reference. A controlled execution experiment shows
2.18--2.23-fold speedups from prefix reuse and cache management, with unchanged
outputs on the tested workload.
A LoRA-adapted Qwen3.5-4B scores 74.39\% versus 66.46\% for a
separately sourced Q8\_0 baseline on the same full-MMLU CUDA workload.
These results support local decision
inference from existing models and adaptation of a smaller model.

\end{abstract}
\section{Introduction}
\label{sec:introduction}
Many applications require a choice among a finite set of answers, such as a
category, a rating, or a control action. A language model can make such a choice
by generating text, but it can also evaluate the candidate answers directly.
Direct evaluation avoids answer parsing and restricts the result to the allowed
set.

Dedicated decision models such as Jev use specialized training for these
tasks~\cite{jev-introduction}. We examine a complementary approach: efficient
decision inference from existing pretrained models, without requiring additional training.
JET (Justification Evaluation in Transformer) scores each answer by its
conditional likelihood and converts the scores into a choice, a binary
probability, or an ordinal estimate.

The contribution is an inference system for local execution. JET shares the
question prefix across candidates and isolates recurrent state during candidate
evaluation. These methods
reduce repeated work while retaining the likelihood-based decision rule.
CPU and GPU placement allow models to run under different memory constraints.
An optional task-aligned LoRA experiment tests whether a smaller model can
improve its decisions without changing the inference interface.

We evaluate decision accuracy and throughput jointly on MMLU and BoolQ using
consumer hardware. The comparison covers model choice and GPU hardware,
and execution optimizations, with Jev as a reference
(Figure~\ref{fig:mmlu-throughput}). Full-MMLU evaluation measures decision
quality and complete-process throughput; a controlled experiment on a fixed
subset measures execution improvements.
Section~\ref{sec:finetuning} reports the separate fine-tuning experiment and
its deployment limitations.
The observed 2.18--2.23-fold speedups preserve outputs in the tested comparison. These results assess the
quality and cost of local decisions; differing evaluation and timing protocols
do not establish parity with Jev.

\section{Related Work}
\label{sec:background}
Pretrained sequence models assign conditional probabilities to token
sequences~\cite{transformer}. Ranking candidate answers by likelihood is an
established evaluation method, supported by tools such as the Language Model
Evaluation Harness~\cite{lm-harness}. Unlike constrained decoding, which limits
the tokens allowed during generation, JET evaluates supplied answers directly.
Its contribution concerns execution of this method rather than a new scoring
objective.

Sharing inference state can reduce repeated computation. PagedAttention enables
cache sharing in language-model serving~\cite{pagedattention}. JET applies prefix
reuse to finite candidate sets and distinguishes attention caches from recurrent
state, which requires isolation between candidate continuations.

Jev targets structured decisions through specialized training and emphasizes
calibrated outputs~\cite{jev-introduction,jev-docs}. JET instead reuses pretrained
model likelihoods, with optional task-aligned adaptation. Both approaches restrict the output space, but valid outputs
do not by themselves imply accurate decisions or calibrated probabilities.

\section{Decision Method}
\label{sec:scoring}
\subsection{Token Probabilities}
Let $x$ be the question context and $\mathcal{C}$ a finite set of candidate
alternatives. The quoted answer text for each $c$ is represented by a token
sequence $y_c=(y_{c,1},\ldots,y_{c,m_c})$, following the model's assistant
prefix. Tokenization preserves the boundary between context and answer.
Let $\mathcal{V}$ be the model vocabulary
and $z_{c,t,v}$ the logit for token $v$ given $(x,y_{c,<t})$.
The probability of each supplied candidate token is
\begin{equation}
  p_{c,t}=p_\theta(y_{c,t}\mid x,y_{c,<t})
  =\frac{\exp(z_{c,t,y_{c,t}})}
         {\sum_{v\in\mathcal{V}}\exp(z_{c,t,v})}.
  \label{eq:token-probability}
\end{equation}
The denominator includes the full vocabulary. Tokens are evaluated under
teacher forcing, without sampling, temperature rescaling, or top-$k$/top-$p$
filtering. The answer text, rather than an arbitrary option label, determines
the score.

For numerical stability, the log-probability $\ell_{c,t}=\log p_{c,t}$ is
written as
\begin{equation}
  \ell_{c,t}=z_{c,t,y_{c,t}}-a_{c,t}
  -\log\sum_{v\in\mathcal{V}}\exp(z_{c,t,v}-a_{c,t}),
  \qquad a_{c,t}=\max_{v\in\mathcal{V}}z_{c,t,v}.
  \label{eq:token-log-probability}
\end{equation}

\subsection{Candidate Likelihoods}
The autoregressive chain rule gives the sequence likelihood and its logarithm:
\begin{equation}
  L_c=p_\theta(y_c\mid x)=\prod_{t=1}^{m_c}p_{c,t},
  \qquad
  s(c\mid x)=\log L_c=\sum_{t=1}^{m_c}\ell_{c,t}.
  \label{eq:score}
\end{equation}
Normalizing these likelihoods over the candidate set gives
\begin{equation}
  q(c\mid x,\mathcal{C})
  =\frac{L_c}{\sum_{d\in\mathcal{C}}L_d}
  =\frac{\exp(s(c\mid x)-M)}
        {\sum_{d\in\mathcal{C}}\exp(s(d\mid x)-M)},
  \quad M=\max_{d\in\mathcal{C}}s(d\mid x).
  \label{eq:probability}
\end{equation}
Equations~\ref{eq:token-probability} and~\ref{eq:probability} normalize over
different sets: vocabulary tokens and complete candidate answers, respectively.
For example, candidate likelihoods of $0.06$ and $0.02$ give normalized
probabilities of $0.75$ and $0.25$.
This distribution is relative to the supplied answers. Scores omit end-of-turn
tokens and are not length-normalized or summed over alternative phrasings.
Answer wording, token length, and the candidate set therefore affect the result.
Adding or duplicating a candidate also changes the normalized distribution.
Normalization alone does not establish probability calibration.

\subsection{Decision Rules}
For a categorical decision, JET selects
\begin{equation}
  \hat c=\operatorname*{arg\,max}_{c\in\mathcal{C}}q(c\mid x,\mathcal{C})
  =\operatorname*{arg\,max}_{c\in\mathcal{C}}s(c\mid x),
  \label{eq:decision}
\end{equation}
with ties resolved by a fixed ordering. A binary question uses
$\mathcal{C}=\{\mathrm{false},\mathrm{true}\}$ and returns
$q(\mathrm{true}\mid x,\mathcal{C})$. An ordinal question with $K$ ordered
levels returns the expected index,
$\sum_{i=0}^{K-1} i\,q(i\mid x,\mathcal{C})$, which may be fractional.
These rules require no additional classification head or task-specific training.

Candidate labels are distinct from answer text. Different labels with identical
text receive the same likelihood and use the fixed tie-breaking rule; binary
and ordinal answer descriptions must be distinct. In evaluation, any label with
the gold answer text is accepted, and its probability is summed with those of
equivalent labels when computing gold negative log-likelihood.

\subsection{Optional Reasoning and Determinism}
JET can generate a reasoning trace once per question and include it in the
context shared by all candidates. Candidate evaluation remains fixed conditional
on that trace, but the complete procedure then includes sampling. The main
execution comparisons disable reasoning generation.

Without reasoning generation, the decision rule is deterministic for fixed
model outputs. Floating-point differences across devices and execution settings
can still change probabilities and decisions. Section~\ref{sec:repeatability}
reports the observed agreement; it does not imply bitwise reproducibility across
hardware.

\section{Efficient Inference}
\label{sec:implementation}
\begin{figure}[t]
\centering
\begin{tikzpicture}[
  box/.style={draw=black!65, rounded corners=2pt, fill=black!3,
    align=center, font=\footnotesize, minimum height=0.85cm, text width=8.0cm},
  branch/.style={box,text width=6.0cm,minimum height=1.1cm},
  flow/.style={-{Stealth[length=2mm]},thick,draw=black!70},
  note/.style={font=\scriptsize,align=center,fill=white,inner sep=2pt}
]
\node[box] (request) at (0,0)
  {\textbf{Request validation}\\State, typed questions, candidates, optional images};
\node[box] (prepare) at (0,-2.0)
  {\textbf{Context and candidate preparation}\\Semantic rendering, chat template, tokenization\\Optional image encoding and visual projection};
\node[box] (prefix) at (0,-4.0)
  {\textbf{Shared-prefix prefill}\\Retain question state; score candidate first tokens};
\node[branch] (kv) at (-3.5,-6.2)
  {\textbf{KV-cache text models}\\Fork prefix state; batch candidate probes\\Partition into waves when required};
\node[branch] (serial) at (3.5,-6.2)
  {\textbf{Hybrid/recurrent or vision path}\\Preserve prefix; reset working sequence\\Evaluate candidate continuations serially};
\node[box] (score) at (0,-8.5)
  {\textbf{Likelihood aggregation}\\Extract target-token log-probabilities\\Sum per candidate; normalize over candidates};
\node[box] (answer) at (0,-10.5)
  {\textbf{Typed response}\\Choice selection or probability aggregation\\Probabilities, usage, and sequence cleanup};
\draw[flow] (request) -- (prepare);
\draw[flow] (prepare) -- (prefix);
\draw[flow] (prefix.south) -- ++(0,-0.32) -| (kv.north);
\draw[flow] (prefix.south) -- ++(0,-0.32) -| (serial.north);
\draw[flow] (kv.south) -- ++(0,-0.35) -| (score.north);
\draw[flow] (serial.south) -- ++(0,-0.35) -| (score.north);
\draw[flow] (score) -- (answer);
\end{tikzpicture}
\caption{End-to-end sampler-free workflow, executed locally on CPU, GPU, or a
mixed placement. Candidate tokens are supplied rather than sampled. Optional
text-only reasoning, omitted from the principal path above, generates a trace
before scoring and freezes it into the shared prefix; that mode includes
sampling. Invalid requests produce errors.}
\label{fig:workflow}
\end{figure}
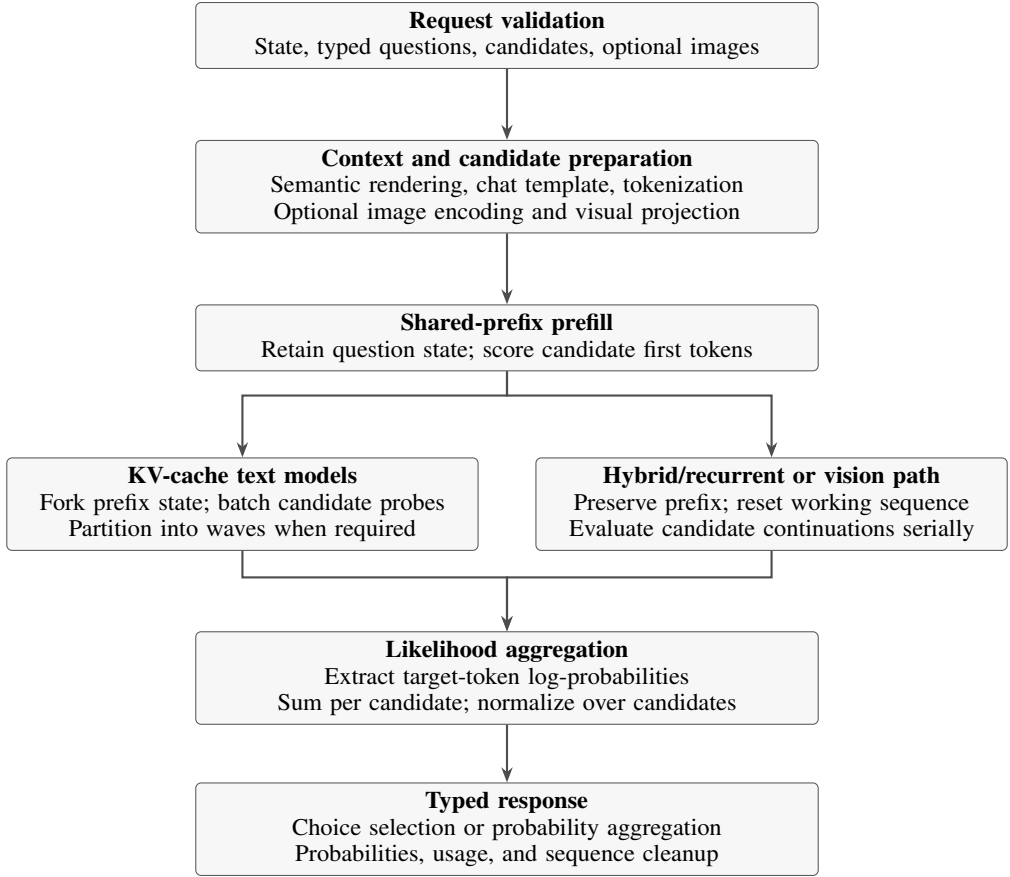

JET implements candidate scoring using llama.cpp~\cite{llamacpp}.
Figure~\ref{fig:workflow} summarizes the decision process. The main efficiency
measures are shared-prefix evaluation and reduced data transfer.

\subsection{Shared-Prefix Evaluation}
For $K$ candidates sharing a prefix of $P$ tokens, independent evaluation
processes $KP$ prefix tokens. Reusing the prefix state reduces this count to
$P$ when all candidates can share one evaluation. The reduction in execution
time also depends on candidate lengths, batching, and memory movement.

For attention-only models, candidate continuations can be evaluated in
batches using the shared prefix. Because the candidates are supplied in
advance, all continuation tokens are available for teacher-forced evaluation.
Batch size is limited by available sequence capacity; larger candidate sets
are evaluated in groups without truncating the question. This does not imply
unlimited parallelism or one forward pass for every request size.

Hybrid models also maintain recurrent state. JET evaluates their continuations
sequentially, restoring the unchanged prefix state before each candidate.
This isolates candidates from one another while preserving prefix reuse.
One stored prefix and one working state suffice, regardless of candidate count.
The prefix output directly supplies the first-token probabilities; single-token
candidates require no further model evaluation.

\subsection{Data Transfer and Computation}
GPU scoring computes probabilities over the full vocabulary but transfers only
the log-probabilities needed for the candidates. This reduces data transfer
without changing the scoring rule.

When a mixture-of-experts model exceeds GPU memory, routed experts can execute
on the CPU while attention and other components execute on the GPU.
Expert weights remain in memory; this placement does not stream selected
weights from disk for each token. Reusing CPU workers reduces repeated startup
overhead when execution alternates between CPU and GPU computation.

\subsection{Multimodal Inference}
\label{sec:multimodal}
The same scoring rule can condition textual candidates on both images and text
using a vision-language model and its matching visual projector. Visual
embeddings count toward the context budget. The current implementation requires
reasoning to be disabled.

A Doom controller illustrates this extension. It selects among nine admissible
controls using 160-by-100 RGB screenshots, visible status information, and recent
actions. Hidden enemy positions and map geometry are excluded. Visual decision
accuracy and episode-level control performance have not been quantitatively
evaluated.

\section{Experimental Evaluation}
\label{sec:evaluation}
\subsection{Experimental Setup}
We use MMLU and BoolQ~\cite{boolq,mmlu} to examine three questions: what
decision quality pretrained models provide, whether they can support local
inference on consumer hardware, and how execution optimizations affect cost
at fixed decision quality. The full-MMLU study compares four models on the Arc
A770 and six on the RTX 4090; a fixed subset supports a controlled prefix-reuse study. The principal
comparisons disable reasoning generation. Appendix~\ref{sec:reproducibility}
describes the data and settings.

Accuracy is measured by agreement with the reference answer: choice predictions
use the highest-probability label, and BoolQ thresholds the true probability at 0.5. Gold negative
log-likelihood (NLL) is computed from the normalized candidate distribution.
The zero-shot prompt and semantic scoring protocol differ from standard
leaderboard procedures; the results characterize the evaluated JET configurations.

Complete-process wall time includes initialization, model loading, evaluation,
and shutdown. Reported throughput is the number of requests divided by this
elapsed time. It is distinct from saturated serving throughput; similarly,
amortized time per request is distinct from a latency percentile. Fresh processes
may reuse operating-system and shader caches. Run counts and configuration
changes are reported for each comparison.

\subsection{Local Deployment on Consumer Systems}
\label{sec:local-deployment}
The evaluated configurations span full placement on a 16\,GiB Arc A770 and
an RTX 4090. These configurations establish a practical deployment range
without server-class accelerators.

On full MMLU with Vulkan and full GPU placement on an RTX 4090, Qwen3.5-27B achieves
85.09\% at 2.21 requests/s and Qwen3.6-35B-A3B achieves 82.62\% at
4.17 requests/s. Section~\ref{sec:mmlu-comparison} reports the complete model
comparison. The Arc A770 full-MMLU runs below cover four models that fit entirely in its
16\,GiB of memory. Their accuracy and throughput refer to the same complete
14,042-question workload. The RTX 4090 runs use a separate machine;
the measured rates are not a controlled hardware speedup comparison.

\subsection{Decision Quality and Model Selection}
Table~\ref{tab:models} reports disabled-thinking decision quality on the same
14,042 MMLU requests using Vulkan on the Arc A770. All four models use Q8\_0
weights and full GPU placement.
The scoring formulation is unchanged by prefix reuse. Response-equivalence
checks in Sections~\ref{sec:prefix-results} and~\ref{sec:repeatability}
validate the tested execution change. Runtime effects are evaluated separately.

\begin{table}[htbp]
\centering\small
\caption{Full MMLU test set on Arc A770, 14,042 requests per model, thinking
disabled. Complete-process wall time includes model loading and shutdown.
One run per model, with no repeat-run variance estimate.}
\label{tab:models}
\begin{tabular}{llrrrr}
\toprule
Model & Quantization & Correct & MMLU (\%) & Wall (s) & Req./s \\
\midrule
Qwen3.5-0.8B & Q8\_0 & 4,615 & 32.87 & 3,015.29 & 4.66 \\
Qwen3.5-2B & Q8\_0 & 6,807 & 48.48 & 4,219.81 & 3.33 \\
Qwen3.5-4B & Q8\_0 & 9,334 & 66.47 & 9,832.76 & 1.43 \\
Qwen3.5-9B & Q8\_0 & 10,332 & 73.58 & 13,307.56 & 1.06 \\
\bottomrule
\end{tabular}
\end{table}

Across the four A770 runs, accuracy rises from 32.87\% with 0.8B to 73.58\%
with 9B, while throughput falls from 4.66 to 1.06 requests/s. These full-set
results provide the model-quality estimates used in this paper.

\subsection{Shared-Prefix Execution}
\label{sec:prefix-results}
The controlled comparison evaluates hybrid prefix reuse and metadata-only cache
resets against independent candidate execution on the Arc A770. Both versions
use the same models and scoring procedure.

\begin{table}[htbp]
\centering\small
\caption{Prefix-reuse and cache-management comparison on 541 requests.
The speedup uses the mean of the two new complete-process times.}
\label{tab:prefix}
\begin{tabular}{lrrrr}
\toprule
Q8\_0 model & Before (s) & After runs (s) & Mean (s) & Speedup \\
\midrule
Qwen3.5-0.8B & 242.54 & 108.06 / 109.10 & 108.58 & $2.23\times$ \\
Qwen3.5-2B & 313.34 & 143.41 / 143.44 & 143.43 & $2.18\times$ \\
\bottomrule
\end{tabular}
\end{table}

All 541 response objects, including probabilities and usage, are exactly equal
to the fresh baseline in both new runs for each model. Prefix reuse reduces prefix-token work by 66.03\%, based on instrumented
new-execution counts and schedule-derived baseline counts. The comparison bundles prefix reuse and cache reset changes, and
does not isolate either contribution or measure the speedup of parallel probes.

\subsection{MMLU Comparison with Larger Models and Jev}
\label{sec:mmlu-comparison}
Table~\ref{tab:full-mmlu} reports complete-process measurements on the same
14,042 MMLU test questions using an RTX 4090 and Vulkan. Each model has one
fresh-process run with reasoning disabled. These full-set measurements replace
subset estimates when comparing model quality and end-to-end throughput.

\begin{table}[htbp]
\centering\small
\caption{Full MMLU test set on RTX 4090 with Vulkan, 14,042 requests per model. Wall time
includes model loading and shutdown; throughput is requests divided by wall time.
One run per model, with no repeat-run variance estimate.}
\label{tab:full-mmlu}
\begin{tabular}{llrrrr}
\toprule
JET model & Quantization & Correct & MMLU (\%) & Wall (s) & Req./s \\
\midrule
Qwen3.5-0.8B & Q8\_0 & 4,552 & 32.42 & 1,434.16 & 9.79 \\
Qwen3.5-2B & Q8\_0 & 6,818 & 48.55 & 1,443.47 & 9.73 \\
Qwen3.5-4B & Q8\_0 & 9,323 & 66.39 & 1,986.14 & 7.07 \\
Qwen3.5-9B & Q8\_0 & 10,344 & 73.66 & 2,466.26 & 5.69 \\
Qwen3.5-27B & Q4\_K\_M & 11,948 & 85.09 & 6,353.75 & 2.21 \\
Qwen3.6-35B-A3B & Q4\_K\_M & 11,601 & 82.62 & 3,370.71 & 4.17 \\
\bottomrule
\end{tabular}
\end{table}

\begin{figure}[H]
\centering
\begin{tikzpicture}[x=1cm,y=1cm,font=\footnotesize]
\node[anchor=west,font=\small\bfseries] at (0,0.8) {Model / deployment};
\node[font=\small\bfseries] at (8.3,0.8) {MMLU accuracy (\%)};
\node[font=\small\bfseries] at (13.1,0.8) {Throughput (req/s)};
\node[font=\scriptsize,text=black!65] at (6.3,0.3) {0};
\node[font=\scriptsize,text=black!65] at (7.02,0.3) {20};
\node[font=\scriptsize,text=black!65] at (7.74,0.3) {40};
\node[font=\scriptsize,text=black!65] at (8.46,0.3) {60};
\node[font=\scriptsize,text=black!65] at (9.18,0.3) {80};
\node[font=\scriptsize,text=black!65] at (9.9,0.3) {100};
\node[font=\scriptsize,text=black!65] at (11.5,0.3) {0};
\node[font=\scriptsize,text=black!65] at (12.275,0.3) {4};
\node[font=\scriptsize,text=black!65] at (13.05,0.3) {8};
\node[font=\scriptsize,text=black!65] at (13.825,0.3) {12};
\node[font=\scriptsize,text=black!65] at (14.6,0.3) {16};
\node[anchor=west,font=\footnotesize\bfseries] at (0,-0.2) {JET / Qwen3.5-0.8B};
\node[anchor=west] at (0.1,-0.63) {Arc A770 (Vulkan)};
\draw[black!12] (6.3,-0.63) -- (9.9,-0.63);
\draw[black!12] (11.5,-0.63) -- (14.6,-0.63);
\fill[blue!65!black] (7.4832,-0.63) circle (2pt);
\node[anchor=east] at (10.8,-0.63) {32.87};
\fill[blue!65!black] (12.4023,-0.63) circle (2pt);
\node[anchor=east] at (15.5,-0.63) {4.66};
\node[anchor=west] at (0.1,-1.06) {RTX 4090 (Vulkan)};
\draw[black!12] (6.3,-1.06) -- (9.9,-1.06);
\draw[black!12] (11.5,-1.06) -- (14.6,-1.06);
\fill[teal!75!black] (7.4670,-1.06) circle (2pt);
\node[anchor=east] at (10.8,-1.06) {32.42};
\fill[teal!75!black] (13.3970,-1.06) circle (2pt);
\node[anchor=east] at (15.5,-1.06) {9.79};
\node[anchor=west] at (0.1,-1.49) {RTX 4090 (CUDA)};
\draw[black!12] (6.3,-1.49) -- (9.9,-1.49);
\draw[black!12] (11.5,-1.49) -- (14.6,-1.49);
\fill[violet!75!black] (7.4793,-1.49) circle (2pt);
\node[anchor=east] at (10.8,-1.49) {32.76};
\fill[violet!75!black] (14.2433,-1.49) circle (2pt);
\node[anchor=east] at (15.5,-1.49) {14.16};
\node[anchor=west,font=\footnotesize\bfseries] at (0,-2.15) {JET / Qwen3.5-2B};
\node[anchor=west] at (0.1,-2.58) {Arc A770 (Vulkan)};
\draw[black!12] (6.3,-2.58) -- (9.9,-2.58);
\draw[black!12] (11.5,-2.58) -- (14.6,-2.58);
\fill[blue!65!black] (8.0451,-2.58) circle (2pt);
\node[anchor=east] at (10.8,-2.58) {48.48};
\fill[blue!65!black] (12.1447,-2.58) circle (2pt);
\node[anchor=east] at (15.5,-2.58) {3.33};
\node[anchor=west] at (0.1,-3.01) {RTX 4090 (Vulkan)};
\draw[black!12] (6.3,-3.01) -- (9.9,-3.01);
\draw[black!12] (11.5,-3.01) -- (14.6,-3.01);
\fill[teal!75!black] (8.0480,-3.01) circle (2pt);
\node[anchor=east] at (10.8,-3.01) {48.55};
\fill[teal!75!black] (13.3848,-3.01) circle (2pt);
\node[anchor=east] at (15.5,-3.01) {9.73};
\node[anchor=west] at (0.1,-3.44) {RTX 4090 (CUDA)};
\draw[black!12] (6.3,-3.44) -- (9.9,-3.44);
\draw[black!12] (11.5,-3.44) -- (14.6,-3.44);
\fill[violet!75!black] (8.0433,-3.44) circle (2pt);
\node[anchor=east] at (10.8,-3.44) {48.43};
\fill[violet!75!black] (14.1473,-3.44) circle (2pt);
\node[anchor=east] at (15.5,-3.44) {13.66};
\node[anchor=west,font=\footnotesize\bfseries] at (0,-4.1) {JET / Qwen3.5-4B};
\node[anchor=west] at (0.1,-4.53) {Arc A770 (Vulkan)};
\draw[black!12] (6.3,-4.53) -- (9.9,-4.53);
\draw[black!12] (11.5,-4.53) -- (14.6,-4.53);
\fill[blue!65!black] (8.6930,-4.53) circle (2pt);
\node[anchor=east] at (10.8,-4.53) {66.47};
\fill[blue!65!black] (11.7767,-4.53) circle (2pt);
\node[anchor=east] at (15.5,-4.53) {1.43};
\node[anchor=west] at (0.1,-4.96) {RTX 4090 (Vulkan)};
\draw[black!12] (6.3,-4.96) -- (9.9,-4.96);
\draw[black!12] (11.5,-4.96) -- (14.6,-4.96);
\fill[teal!75!black] (8.6902,-4.96) circle (2pt);
\node[anchor=east] at (10.8,-4.96) {66.39};
\fill[teal!75!black] (12.8698,-4.96) circle (2pt);
\node[anchor=east] at (15.5,-4.96) {7.07};
\node[anchor=west] at (0.1,-5.39) {RTX 4090 (CUDA)};
\draw[black!12] (6.3,-5.39) -- (9.9,-5.39);
\draw[black!12] (11.5,-5.39) -- (14.6,-5.39);
\fill[violet!75!black] (8.6925,-5.39) circle (2pt);
\node[anchor=east] at (10.8,-5.39) {66.46};
\fill[violet!75!black] (13.3968,-5.39) circle (2pt);
\node[anchor=east] at (15.5,-5.39) {9.79};
\node[anchor=west] at (0.1,-5.82) {RTX 4090, LoRA (CUDA)};
\draw[black!12] (6.3,-5.82) -- (9.9,-5.82);
\draw[black!12] (11.5,-5.82) -- (14.6,-5.82);
\draw[green!55!black] (8.9781,-5.82) circle (2.8pt);
\node[anchor=east] at (10.8,-5.82) {74.39};
\draw[green!55!black] (13.4698,-5.82) circle (2.8pt);
\node[anchor=east] at (15.5,-5.82) {10.17};
\node[anchor=west,font=\footnotesize\bfseries] at (0,-6.48) {JET / Qwen3.5-9B};
\node[anchor=west] at (0.1,-6.91) {Arc A770 (Vulkan)};
\draw[black!12] (6.3,-6.91) -- (9.9,-6.91);
\draw[black!12] (11.5,-6.91) -- (14.6,-6.91);
\fill[blue!65!black] (8.9489,-6.91) circle (2pt);
\node[anchor=east] at (10.8,-6.91) {73.58};
\fill[blue!65!black] (11.7044,-6.91) circle (2pt);
\node[anchor=east] at (15.5,-6.91) {1.06};
\node[anchor=west] at (0.1,-7.34) {RTX 4090 (Vulkan)};
\draw[black!12] (6.3,-7.34) -- (9.9,-7.34);
\draw[black!12] (11.5,-7.34) -- (14.6,-7.34);
\fill[teal!75!black] (8.9519,-7.34) circle (2pt);
\node[anchor=east] at (10.8,-7.34) {73.66};
\fill[teal!75!black] (12.6031,-7.34) circle (2pt);
\node[anchor=east] at (15.5,-7.34) {5.69};
\node[anchor=west] at (0.1,-7.77) {RTX 4090 (CUDA)};
\draw[black!12] (6.3,-7.77) -- (9.9,-7.77);
\draw[black!12] (11.5,-7.77) -- (14.6,-7.77);
\fill[violet!75!black] (8.9483,-7.77) circle (2pt);
\node[anchor=east] at (10.8,-7.77) {73.56};
\fill[violet!75!black] (13.1168,-7.77) circle (2pt);
\node[anchor=east] at (15.5,-7.77) {8.34};
\node[anchor=west,font=\footnotesize\bfseries] at (0,-8.43) {JET / Qwen3.5-27B};
\node[anchor=west] at (0.1,-8.86) {RTX 4090 (Vulkan)};
\draw[black!12] (6.3,-8.86) -- (9.9,-8.86);
\draw[black!12] (11.5,-8.86) -- (14.6,-8.86);
\fill[teal!75!black] (9.3632,-8.86) circle (2pt);
\node[anchor=east] at (10.8,-8.86) {85.09};
\fill[teal!75!black] (11.9282,-8.86) circle (2pt);
\node[anchor=east] at (15.5,-8.86) {2.21};
\node[anchor=west] at (0.1,-9.29) {RTX 4090 (CUDA)};
\draw[black!12] (6.3,-9.29) -- (9.9,-9.29);
\draw[black!12] (11.5,-9.29) -- (14.6,-9.29);
\fill[violet!75!black] (9.3573,-9.29) circle (2pt);
\node[anchor=east] at (10.8,-9.29) {84.92};
\fill[violet!75!black] (12.2099,-9.29) circle (2pt);
\node[anchor=east] at (15.5,-9.29) {3.66};
\node[anchor=west,font=\footnotesize\bfseries] at (0,-9.95) {JET / Qwen3.6-35B-A3B};
\node[anchor=west] at (0.1,-10.38) {RTX 4090 (Vulkan)};
\draw[black!12] (6.3,-10.38) -- (9.9,-10.38);
\draw[black!12] (11.5,-10.38) -- (14.6,-10.38);
\fill[teal!75!black] (9.2742,-10.38) circle (2pt);
\node[anchor=east] at (10.8,-10.38) {82.62};
\fill[teal!75!black] (12.3071,-10.38) circle (2pt);
\node[anchor=east] at (15.5,-10.38) {4.17};
\node[anchor=west] at (0.1,-10.81) {RTX 4090 (CUDA)};
\draw[black!12] (6.3,-10.81) -- (9.9,-10.81);
\draw[black!12] (11.5,-10.81) -- (14.6,-10.81);
\fill[violet!75!black] (9.2727,-10.81) circle (2pt);
\node[anchor=east] at (10.8,-10.81) {82.57};
\fill[violet!75!black] (12.4616,-10.81) circle (2pt);
\node[anchor=east] at (15.5,-10.81) {4.96};
\node[anchor=west,font=\footnotesize\bfseries] at (0,-11.47) {Jev 1.13};
\node[anchor=west] at (0.1,-11.9) {Hosted API};
\draw[black!12] (6.3,-11.9) -- (9.9,-11.9);
\draw[black!12] (11.5,-11.9) -- (14.6,-11.9);
\fill[orange!80!black] (9.5062,-11.9) circle (2pt);
\node[anchor=east] at (10.8,-11.9) {89.06};
\fill[orange!80!black] (12.0541,-11.9) circle (2pt);
\node[anchor=east] at (15.5,-11.9) {2.86};
\end{tikzpicture}
\caption{Full-MMLU accuracy and throughput with reasoning disabled. Blue: Arc A770 (Vulkan); teal: RTX 4090 (Vulkan); violet: RTX 4090 (CUDA); green outlined: Qwen3.5-4B LoRA (CUDA); orange: Jev 1.13 hosted API. Each JET point combines accuracy and complete-process throughput from the same 14,042-question run. Jev uses full-MMLU accuracy and a separately reported API rate. Each metric uses a common horizontal scale.}
\label{fig:mmlu-throughput}
\end{figure}
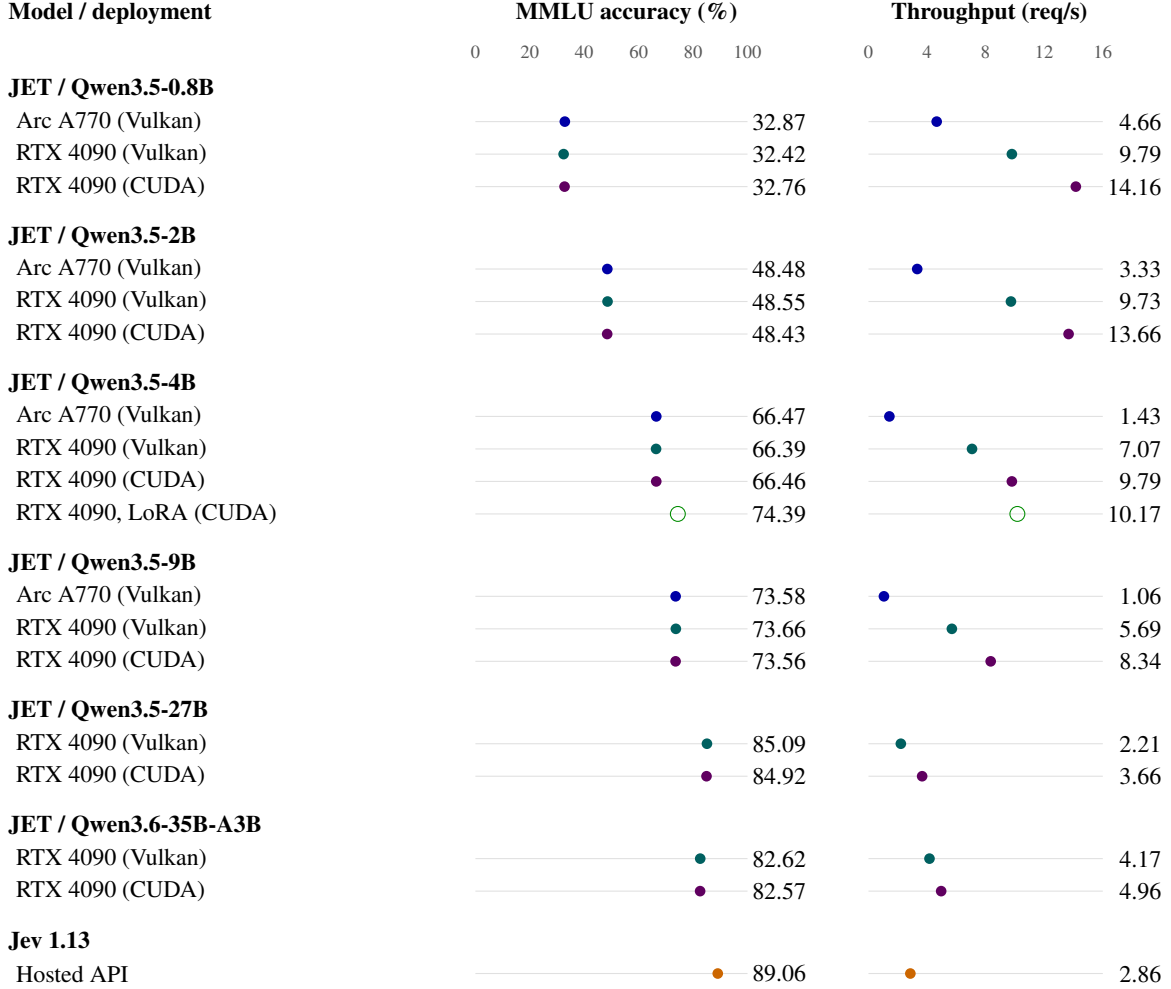

Qwen3.5-27B answers 347 more questions correctly than Qwen3.6-35B-A3B,
while the latter processes approximately 1.89 times as many requests per second.
The architectures and model generations differ, so this contrast does not
isolate the effect of parameter count. Figure~\ref{fig:mmlu-throughput} adds
corresponding full-set CUDA runs of the same six models on the RTX 4090. The
Qwen3.6-35B-A3B CUDA run scores 11,595/14,042 (82.57\%) in 2,829.36 seconds
(4.96 requests/s); Qwen3.5-27B scores 11,925/14,042 (84.92\%) in
3,832.36 seconds (3.66 requests/s). For Qwen3.6, 328 top-1 choices differ
between the full-set CUDA and Vulkan runs. A same-executable backend check on
a fixed subset confirms that switching backend changes decisions
(Appendix~\ref{sec:backend-sensitivity}). Table~\ref{tab:full-mmlu}
retains the Vulkan measurements.

Table~\ref{tab:mmlu-comparison} places selected JET results alongside a manually
measured Jev score and published results for larger instruction-tuned models.
On the complete MMLU test set, JET with Qwen3.5-27B and Vulkan on an RTX 4090
achieves 85.09\%. The author's manual evaluation of Jev 1.13 yields 89.06\%,
with a separately reported API throughput of 2.86 requests/s. Jev's evaluated
interface is sampler-free and exposes no configurable reasoning or few-shot
settings. These Jev results are the author's measurements. Published
larger-model results retain their source evaluation protocols.

\begin{table}[htbp]
\centering\small
\caption{MMLU reference comparison. Protocols differ across groups; the table
provides numerical context, not a matched ranking. External results are not
measurements of those models running in JET.}
\label{tab:mmlu-comparison}
\begin{tabular}{>{\raggedright\arraybackslash}p{6.0cm}r>{\raggedright\arraybackslash}p{6.2cm}}
\toprule
System / model & MMLU (\%) & Evaluation basis \\
\midrule
\multicolumn{3}{l}{\emph{Full-MMLU evaluations}} \\
JET / Qwen3.5-27B, RTX 4090 & 85.09 & Full MMLU test set; Vulkan; local execution \\
JET / Qwen3.6-35B-A3B, RTX 4090 & 82.62 & Full MMLU test set; Vulkan; local execution \\
Jev 1.13 & 89.06 & Full MMLU test set; native decision interface \\
\midrule
\multicolumn{3}{l}{\emph{Published larger-model references}} \\
Llama 3.1 70B Instruct~\cite{llama31-card} & 83.6 & 5-shot; subject-macro accuracy \\
Llama 3.1 405B Instruct~\cite{llama31-card} & 87.3 & 5-shot; subject-macro accuracy \\
Qwen2.5-72B-Instruct~\cite{deepseek-v3-results} & 85.3 & MMLU exact match in the cited comparison \\
DeepSeek-V3 (671B)~\cite{deepseek-v3-results} & 88.5 & MMLU exact match in the cited comparison \\
\bottomrule
\end{tabular}
\end{table}

The outlined Qwen3.5-4B point is the separately evaluated LoRA adaptation;
Section~\ref{sec:finetuning} describes its training and comparison boundary.

The highest JET full-MMLU score in this study is 3.97 percentage points below
the author's Jev 1.13 result. Shared benchmark coverage does
not ensure identical prompting, answer scoring, or aggregation, so the difference
alone does not establish statistical equivalence. Published larger-model scores
provide additional context under their respective evaluation protocols; they do
not establish how those models would perform under JET's candidate-scoring method.

\paragraph{Functional and performance comparison with Jev.}
Both systems provide sampler-free decisions over supplied alternatives,
including boolean probabilities, named choices, and rubric scores. JET obtains
this behavior from reusable pretrained models with local execution. JET
computes its score as the expectation of zero-based level indices and returns
candidate probability maps for choice and score questions. Jev additionally
reports separate confidence values for those question types; JET's normalized
likelihoods are not presented as equivalent calibrated confidence estimates. Jev describes
parallel evaluation of independent questions; JET's hybrid path evaluates
candidate continuations serially while sharing their prefix~\cite{jev-docs,jev-workflows}.

The manual Jev 1.13 measurement reports 2.86 requests/s via API. On full MMLU
with Vulkan, JET reaches 4.17 requests/s with Qwen3.6-35B-A3B or 2.21 requests/s with
Qwen3.5-27B on the RTX 4090, including startup and model loading. The reported
rates show a local accuracy--throughput choice, but differences in timing
boundaries preclude a controlled speedup claim. JET exposes local placement
and model choice; the evaluated Jev interface does not
expose a reasoning mode. Its hardware configuration is unknown.

Figure~\ref{fig:mmlu-throughput} summarizes evaluated
JET configurations for each model and GPU backend, with reasoning
disabled, alongside Jev. Early prompt baselines, intermediate optimization
variants, and repeated measurements are omitted from the figure. Model choice changes decision quality; the prefix
comparison improves throughput without changing responses. Within the RTX 4090 Vulkan full-set comparison, Qwen3.5-27B provides the
highest MMLU accuracy, while Qwen3.6-35B-A3B provides greater throughput.
These results distinguish gains from model capability and execution efficiency.

Each Arc A770 and RTX 4090 point shows accuracy and throughput from the
same full-MMLU run. Jev uses
full-MMLU accuracy and a separately reported API rate.
Published larger-model scores remain in Table~\ref{tab:mmlu-comparison}
because corresponding throughput measurements are unavailable.

\subsection{Numerical Equivalence and Reproducibility}
\label{sec:repeatability}
The prefix comparison preserves exact responses
in its tested configurations. This does not imply agreement across platforms.

Independent candidate execution provides a reference for testing state
isolation. An initial hybrid schedule changed 9 of 20 diagnostic answers.
Restoring independent candidate states recovered agreement on all top-1
decisions in that slice, with a maximum probability difference of
$1.12\times10^{-7}$. A separate experiment batching independent requests
exceeded the same numerical tolerance and was not retained. These checks
support the use of serial candidate continuations in the reported hybrid
experiments.

\section{Task-Aligned Fine-Tuning}
\label{sec:finetuning}
We fine-tuned Qwen3.5-4B with LoRA~\cite{lora} to test whether training on
JET's decision format improves accuracy. Training updates a small set of
parameters while keeping the pretrained weights frozen. The candidate-scoring
method remains unchanged.

\subsection{Training Method}
The training mixture contains 22,791 records: 9,427 BoolQ~\cite{boolq},
2,248 ARC-Easy and 1,116 ARC-Challenge~\cite{arc}, and 10,000
SNLI~\cite{snli}. Each example uses the same prompt and answer format as
JET inference, with reasoning disabled. Training minimizes cross-entropy over
the correct answer text, matching the continuation scored at inference.

We train 32.46 million parameters (0.77\% of the model) for one epoch, using
LoRA rank 16, a learning rate of $10^{-4}$, and an effective batch size of 16.
A separate 512-example validation subset monitors loss. Official test splits
are held out, duplicate examples are removed, and exact matches to the MMLU
evaluation questions are excluded. MMLU is not used for training or model
selection, although these checks cannot rule out paraphrases or pretraining
contamination. Full training settings are provided in the repository.

\subsection{Measured Improvement}
Both models are evaluated with Q8\_0 quantization on the same 14,042 MMLU
questions using CUDA on the RTX 4090 and the same JET settings. Accuracy rises
from 66.46\% to 74.39\%, a gain of 7.93 percentage points. Mean gold NLL
falls from 0.9671 to 0.7721. The adapted model corrects 1,770 answers and
changes 656 correct answers to incorrect ones; accuracy improves in 53 of 57
subjects. Figure~\ref{fig:mmlu-throughput} compares the adapted model with the
other evaluated models.

Complete-process throughput is 9.79 requests/s for the reference model and
10.17 requests/s for the adapted model. With one run per model, this difference
does not establish a repeatable speedup.

\subsection{Limitations}
The reference model comes from a separate conversion, and we have not verified
that it uses exactly the same pretrained weights as the training checkpoint.
The accuracy gain therefore cannot be attributed entirely to LoRA. A controlled
comparison would evaluate the same checkpoint with and without adaptation,
using the same conversion procedure.

This experiment covers text-based choices only. It does not evaluate adaptation
for visual inputs, ordinal scores, or decisions in closed-loop applications.
The MMLU improvement may not generalize to those tasks.

\section{Discussion and Limitations}
\label{sec:limitations}
The results support local decision inference on the tested GPU
configurations. Figure~\ref{fig:mmlu-throughput} summarizes accuracy and
throughput across the selected models and deployments. The separate execution
studies show that optimizations can raise throughput at unchanged accuracy.
Full GPU placement supports the measured models on the two devices.
Performance still depends on model size, candidate count and length, and hardware.

The model comparison evaluates the complete MMLU test set but has one run per
model and backend, so it provides no repeat-run variance estimate. Model versions,
quantization, placement, and timing boundaries differ across studies, so their
results cannot be treated as a controlled hardware comparison. The prefix study
changes both state reuse and cache management; it does not isolate their
individual effects. Shared benchmark coverage also does not make JET and Jev
directly comparable: prompts, scoring, aggregation, and local versus API timing
may differ. A matched constrained-generation baseline is needed to separate the
inference method from the underlying model's capability.
Comparison with a fixed Jev version should use the same contexts, candidate
meanings, answer types, and evaluation rules. Performance measurements should
report warm p50/p95 latency, throughput at fixed concurrency, startup cost,
and failure rates, distinguishing local inference from remote service latency.

Candidate likelihoods are sensitive to answer wording, length, and answer priors. Their
normalization does not establish calibration~\cite{calibration}; this requires
separate empirical evaluation. NLL is a diagnostic, not a substitute for
calibration curves, Brier scores, or selective-prediction evaluation.
Pretraining data may also overlap benchmark content.
The fine-tuning experiment excludes the frozen MMLU requests and exact
normalized overlaps, but cannot rule out paraphrases or pretraining
contamination. Its public text tasks do not establish improved visual,
ordinal, or closed-loop decision quality. The training mixture, adapter rank,
and quantization have no ablation study.

Sampler-free selection removes sampling variance, but numerical differences
across hardware, quantization, kernels, and batch layouts can remain. Exact
agreement applies only to the tested configurations. Hybrid continuations are
currently evaluated serially; concurrency across independent questions requires
state isolation and numerical validation. Multimodal accuracy and closed-loop
control remain unmeasured. Raw experiment artifacts are retained outside version
control, so independent replication requires those records or regeneration under
the documented conditions.

\section{Conclusion}
\label{sec:conclusion}
JET applies pretrained language and vision-language models to finite decisions
through candidate likelihood evaluation without requiring additional training.
The accuracy--throughput comparison in Figure~\ref{fig:mmlu-throughput} shows
the range of local deployment choices with full GPU placement on an Arc A770
and an RTX 4090. In the six-model full-MMLU Vulkan comparison on an RTX 4090,
Qwen3.5-27B has the highest accuracy, 85.09\% at 2.21 requests/s.
Qwen3.6-35B-A3B reaches 82.62\% at 4.17 requests/s, offering 1.89 times
the throughput at 2.47 percentage points lower accuracy.
In the corresponding CUDA series, Qwen3.5-27B reaches 84.92\% at
3.66 requests/s and Qwen3.6-35B-A3B reaches 82.57\% at 4.96 requests/s.
Jev provides a reference for decision quality and
request rate, although differences in evaluation and timing prevent a controlled
performance ranking.
The LoRA-adapted Qwen3.5-4B scores 74.39\% on full-MMLU CUDA versus 66.46\%
for a separately sourced Q8\_0 baseline under the same JET scoring protocol.
The differing source/conversion provenance and lack of repeat runs prevent a
LoRA-only causal or speed-equivalence claim. Application-specific holdouts
remain necessary.

The 2.18--2.23-fold speedups from prefix reuse and cache management show that
this execution optimization reduces cost while preserving outputs in the tested
configurations. Future work
should measure calibration, latency distributions, and
multimodal decision quality, and compare decision systems under matched conditions.

\section*{Statements}
\subsection*{AI-Assisted Writing}
The author used AI-assisted writing tools for language editing, organization, and
drafting support during preparation of this manuscript. The author reviewed and
edited the resulting text and takes full responsibility for the content, claims,
experiments, and conclusions.

\bibliographystyle{unsrtnat}
\bibliography{references}
\clearpage
\appendix
\section{Experimental Protocol}
\label{sec:reproducibility}
Source code and reproduction instructions are available at
\url{https://github.com/yet-another-ai/jet}.
The \href{https://github.com/yet-another-ai/jet#build}{installation instructions}
and \href{https://github.com/yet-another-ai/jet/blob/main/tests/accuracy/README.md}{evaluation guide}
cover setup, data and model preparation, and benchmark execution.
Detailed configurations and recorded measurements are provided in the
\href{https://github.com/yet-another-ai/jet/blob/main/docs/accuracy.md}{accuracy notes}
and \href{https://github.com/yet-another-ai/jet/blob/main/docs/performance.md}{performance notes}.

The primary experiments were conducted on September 21--29, 2026; the Arc A770
full-set study ran on September 28--29. An initial CUDA check ran on September 28, followed by the
remaining RTX 4090 CUDA full-set runs on September 29.
The sampled workload comprises 256 label-balanced BoolQ validation examples
and five MMLU test examples from each of 57 subjects, totaling 541 requests
(seed 20260921). The Arc A770 and RTX 4090 model comparisons use all
14,042 MMLU test questions across 57 subjects. Data preparation verifies source checksums.
JET uses zero-shot semantic candidate scoring, with reasoning
disabled unless stated otherwise. Models up to Qwen3.5-9B use Q8\_0 weights;
Qwen3.5-27B and Qwen3.6-35B-A3B use Q4\_K\_M. Arc A770 experiments use Vulkan;
RTX 4090 full-set measurements include Vulkan and CUDA. The Qwen3.6 text
benchmark does not load a visual projector.

The Arc A770 full-MMLU study uses an i9-13900K, 64\,GiB RAM, and a 16\,GiB
GPU. Qwen3.5-0.8B, 2B, 4B, and 9B use full GPU placement,
eight CPU threads, two sequence slots, a microbatch size of 256, an output-row
limit of 256, eight requests per chunk, and no memory mapping. Each has one
complete run with 14,042 responses and zero failures. All four use the same
input, gold labels, and evaluation settings.

The full-MMLU study ran on September 23--24, 2026, using the RTX 4090 and
Vulkan. Each of six models had one fresh-process run with 14,042 requests,
eight input requests per chunk, two sequence slots, microbatch size 256,
output-row limit 256, eight CPU threads, and no memory mapping. All runs used
the same request and gold files. Each run produced 14,042 responses and zero
evaluation failures.
The complete input files can be regenerated from the pinned MMLU archive with
\texttt{prepare-accuracy-data.py} using \texttt{--boolq-limit 0 --mmlu-all
--ascii-json}.
The evaluator treats identical answer texts under different MMLU option labels
as equivalent. Per-run response, timing, provenance, and subject-level reports
are retained locally outside version control.
The RTX 4090 CUDA series used the same six models, input, gold, scoring
protocol, placement, and run settings. Qwen3.6-35B-A3B ran on September 28;
the other five ran on September 29. A control using the same executable and
model files for both GPU backends confirms that switching the backend changes
answers on its sampled rows (Appendix~\ref{sec:backend-sensitivity}).

The separate Qwen3.5-4B LoRA run finished September 30, 2026. Its pinned
base-model revision, source-file checksums, data split manifest, optimizer
configuration, adapter, and export provenance are recorded under
\texttt{training/}. The MMLU comparison uses the September 29 CUDA base run
and the September 30 adapted run, with identical input/gold hashes, executable
hash, backend, batch size, and JET flags. Both produce 14,042 responses and zero
failures. The adapted model is a merged text-only Q8\_0 GGUF made with
\texttt{--no-nextn}; the pre-existing base GGUF includes MTP weights. Each
configuration has one complete-process timing run. The paired analysis counts
semantic answer equivalence when multiple option labels share the same text.
The base GGUF was obtained from a separate Q8\_0 conversion repository; a
zero-adapter GGUF export of the pinned training checkpoint has not been
evaluated, so the comparison does not isolate LoRA from source provenance.

The shared-prefix comparison uses a preserved baseline executable. Both
versions use eight input requests per chunk, nine sequence slots, 2,048 context
tokens per sequence, token batches of 2,048, a microbatch size of 512, and an
output-row limit of 256. Each model has one baseline run and two optimized runs,
with model order reversed in the second round and no explicit warm-up.
Optimized execution uses 541 prefix evaluations and 128,583 prefix tokens;
the baseline schedule implies 1,652 prefix evaluations and 378,526 prefix
tokens. Continuations comprise 9,795 tokens across 2,369 decoder calls.

Reported JET timings cover complete processes, including model loading.
The full-MMLU rates in Tables~\ref{tab:models} and~\ref{tab:full-mmlu} are measured on the same
requests as their corresponding accuracy scores.
Jev 1.13's manually measured 89.06\% full-MMLU accuracy and 2.86 requests/s API
throughput lack documented concurrency and detailed timing boundaries. Its
interface does not expose configurable reasoning or few-shot settings.

Raw artifacts are excluded from the repository. Prefix
records were checked locally. The full-MMLU records include local responses,
reports, and provenance.

\subsection{GPU Backend Sensitivity}
\label{sec:backend-sensitivity}
Even on the same GPU, CUDA and Vulkan can produce different top-1 answers.
On a fixed 570-question RTX 4090 subset, changing only the backend with one
executable changes 16, 12, and 14 choices for Qwen3.5-0.8B, Qwen3.5-2B,
and Qwen3.6-35B-A3B, respectively. Current Vulkan outputs for all three
models match the historical full-run responses projected onto this subset,
including probabilities; the Qwen3.6 CUDA outputs also match its historical
full-run projection. These checks link the controlled comparison to the
reported full-set runs on the sampled rows.
Changes from correct to incorrect and from incorrect to correct often offset
one another in aggregate accuracy, masking differences on individual questions.
Vulkan execution options also affect answers: disabling
\texttt{NV\_coopmat2}, integer dot product, or FP16 support separately changes
results, yet none of these single-option changes makes Vulkan fully agree
with CUDA or CPU execution. The observed Vulkan differences between the
Arc A770 and RTX 4090 arise primarily because the A770 cannot enable
\texttt{KHR\_cooperative\_matrix}. Consequently, small accuracy differences
across devices or backends should not be read as isolated model-quality changes;
paired per-question responses are needed to reveal the underlying changes.

\end{document}